\pdfoutput=1

\documentclass[11pt]{article}

\usepackage[]{ACL2023}

\usepackage{color}

\usepackage{times}
\usepackage{multirow}
\usepackage{multicol}
\usepackage{graphicx}
\usepackage{booktabs}
\usepackage{adjustbox}
\usepackage{tabularx}
\usepackage{latexsym}
\usepackage{booktabs}
\usepackage{siunitx}
\usepackage{amsmath}
\usepackage{mathtools}
\usepackage{xspace}
\usepackage[T1]{fontenc}

\usepackage[utf8]{inputenc}

\usepackage{microtype}
\usepackage{booktabs} 
\usepackage{array}    
\usepackage{inconsolata}
\newcommand{\modelname}{FactDetect\xspace}

\usepackage{titlesec}
\titlespacing*{\section}{0pt}{0.1\baselineskip}{0.2\baselineskip}

\title{Robust Claim Verification Through Fact Detection}

\author{Nazanin Jafari \\
  University of Massachusetts Amherst \\
  Amherst, MA, USA \\
  \texttt{nazaninjafar@cs.umass.edu} \\\And
  James Allan \\
  University of Massachusetts Amherst \\
  Amherst, MA, USA \\
  \texttt{allan@cs.umass.edu} \\}

\begin{document}
\maketitle
\begin{abstract}

Claim verification can be a challenging task. In this paper, we present a method to enhance the robustness and reasoning capabilities of automated claim verification through the extraction of short facts from evidence. Our novel approach, \modelname, leverages Large Language Models (LLMs) to generate concise factual statements from evidence and label these facts based on their semantic relevance to the claim and evidence. The generated facts are then combined with the claim and evidence. To train a lightweight supervised model, we incorporate a fact-detection task into the claim verification process as a multitasking approach to improve both performance and explainability. We also show that augmenting \modelname in the claim verification prompt enhances performance in zero-shot claim verification using LLMs.

Our method demonstrates competitive results in the supervised claim verification model by $15\%$ on the F1 score when evaluated for challenging scientific claim verification datasets. We also demonstrate that \modelname can be augmented with claim and evidence for zero-shot prompting (Aug\modelname) in LLMs for verdict prediction. We show that Aug\modelname outperforms the baseline with statistical significance on three challenging scientific claim verification datasets with an average of $17.3\%$ performance gain compared to the best performing baselines.
\end{abstract}

\begin{figure}[t]
    \centering
    \includegraphics[width = 0.85\columnwidth]{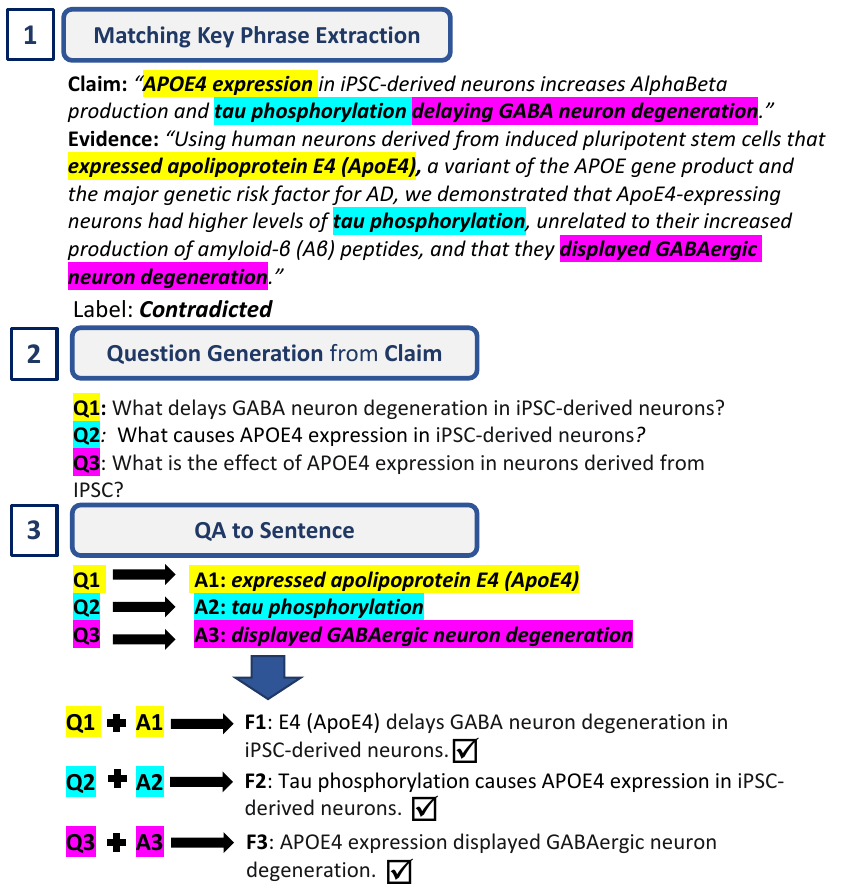}
    \caption{\small{Three-step process of short fact generation from evidence. 1) First we use LLM to generate matching phrases between claim and evidence. 2) Using the extracted phrases from \textbf{claim} we design a question generation to generate questions from the claim and the given phrase. 3) The generated matching phrase from \textbf{evidence} is concatenated with the question generated from \textbf{claim} for short fact generation. Check marks suggest the importance of generated sentences.}}\label{fig:fact_gen}
\end{figure}
\section{Introduction}

Due to the proliferation of disinformation in many online platforms such as social media, automated claim verification has become an important task in natural language processing (NLP). ``Claim verification'' refers to predicting the verdict for a claim -- is it supported or contradicted by a piece of evidence that has been extracted from a corpus of documents ~\cite{Thorne18Fever,wadden-etal-2022-scifact,guo2022survey}.

Claim verification can be challenging for several reasons. First, the available human-annotated data is limited, resulting in limited performance by current trained models. The task is even harder for scientific claim verification where the claim and the corresponding evidence belong to specific scientific domains, generally requiring specialized knowledge of scientific background, numerical reasoning, and statistics ~\cite{wadden-etal-2020-fact}. A key challenge in developing automated claim verification systems lies in accurately representing the subtleties of the task. This includes the capacity to change a verdict from `supported' to change a verdict from `supported' to `contradicted' when new evidence in the test set contradicts what was in the training set.

Human-based reasoning for this task involves creating a meaningful link between the claim and the evidence and performing reasoning on such links. A few studies have proposed reasoning methods based on question answering \cite{pan-etal-2021-Zero-shot-FV, dai2022ask, lee2021crossaug}, and more recent approaches leverage Large Language Models (LLMs) to generate reasoning programs~\cite{pan-etal-2023-fact} or decompose claims into first-order logic clauses~\cite{wang2023explainable}. Question-answering, which involves asking questions about the claim or evidence, retrieving answers from each component, and using these answers for subsequent tasks, is one method used to improve reasoning and explanation in claim verification tasks ~\cite{pan-etal-2021-Zero-shot-FV, dai2022ask}. Intuitively, a question asked about a supported or contradicted claim should be \emph{answerable} by the corresponding evidence. The evidence-provided answer can offer critical factual information for veracity prediction.

Motivated by these reasoning approaches, we introduce \modelname. This short sentence generation framework enhances the state-of-the-art trained models and LLMs by simplifying the connection between claim and evidence pairs by identifying and distilling crucial facts from evidence and then transforming these facts into simpler and concise sentences. We hypothesize that these concise sentences will enhance reasoning abilities by including scientific understanding, simplifying the connection between a claim and its complex scientific evidence, and making a meaningful connection between the claim and the evidence.
\modelname comprises: a) short fact generation 
b) weakly labeling the short facts based on their importance given the claim; and, c) using these facts in either a multi-task learning-based training of a supervised claim verification model or as an extra step to improve the performance of zero-shot claim-verification using LLMs. 
An overview of the fact-generation process with an example is given in Figure~\ref{fig:fact_gen}.

We evaluate \modelname in either multi-task-based finetuning of claim verification models or zero-shot claim verification through LLMs on three scientific claim-verification datasets: SciFact~\cite{wadden-etal-2020-fact}, HealthVer~\cite{sarrouti2021evidence} and Scifact-Open~\cite{wadden-etal-2022-scifact}. 

\begin{figure*}[ht]
\centering    \includegraphics[width=0.9\textwidth]{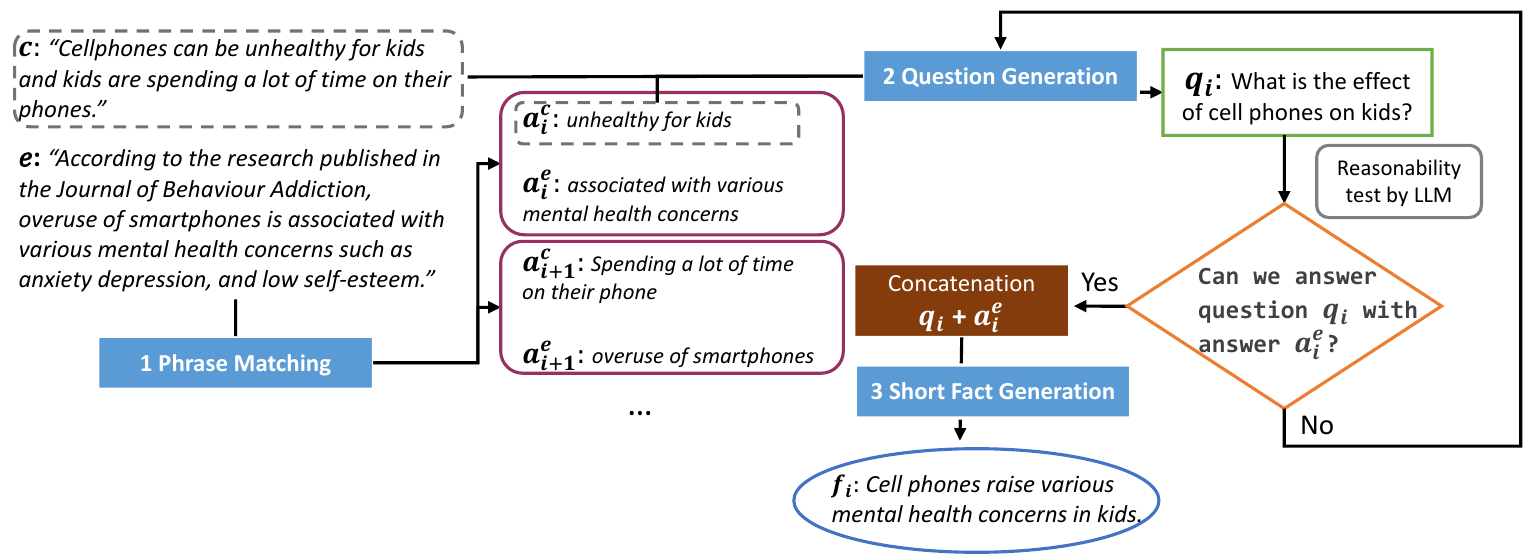}
    \caption{\small{Overview of the proposed framework. \modelname consists of three steps of 1) Phrase matching, 2) Question generation and finally 3) Short fact generation. }  }
    \label{fig:factgen}
\end{figure*}

In summary, our contributions are: 1) an effective approach for decomposing evidence sentences into shorter sentences. Our method prioritizes relevance to the claim and importance for the verdict, based on the connection between evidence and the claim. 2) \modelname enhances the performance of supervised claim verification models in the proposed multi-task learning model. 3) augmenting \modelname generated short sentences for relevant fact detection and claim verification demonstrates state-of-the-art performance in the majority of the LLMs in the few-shot prompting setting.     
The code and data are available at https://https://github.com/nazaninjafar/factdetect.

\section{Background}
Automated claim verification means determining the veracity of a claim, typically by retrieving likely relevant documents and searching for evidence within them. The key objective is to ascertain if the evidence either \emph{supports}, \emph{contradicts} or does not have \emph{enough information} to verify the claim. Various datasets have been proposed to facilitate research in this area in different domains: e.g., FEVER ~\cite{Thorne18Fever} is a Wikipedia-based claim verification dataset. Claim verification in the scientific setting has also been proposed in recent years to facilitate research in this complex domain~\cite{wadden-etal-2022-scifact, wadden-etal-2020-fact, saakyan2021covid,sarrouti2021evidence,kotonya-toni-2020-explainable-automated,diggelmann2020climatefever}. The datasets used for these problems, despite their value, often have limited training data due to the high cost of creation, impacting the reasoning capabilities and robustness of claim verification methods.

In addressing these challenges, the literature shows significant advances in models for verifying scientific claims through reasoning. Prior studies have explored using attention mechanisms to identify key evidence segments \cite{popat2017truth, 10.1145/3357384.3357862, 10.1145/3308558.3314119,info13100500}. Recently, the integration of LLMs in explanation generation has been investigated. For example, ProofVer~\cite{krishna2022proofver} generates proofs for the claim based on evidence using logic-based inference.
ProgramFC ~\cite{pan-etal-2023-fact} uses LLMs to generate reasoning programs that can be used to guide fact-checking, and FOLK~\cite{wang2023explainable} leverages the in-context learning ability of LLMs to generate First Order Logic-Guided reasoning over a set
of knowledge-grounded question-and-answer
pairs to make veracity predictions without using annotated evidence. Other sets of studies attempt to improve this problem through sentence simplification and evidence summarization using LLMs (e.g., ~\cite{mehta2022improving,stammbach2020fever}).

Our work diverges from these methods as we propose an add-on task to enhance the robustness and reasoning ability of existing models. This is achieved through a novel data augmentation strategy which improves the connection between claims and evidence by focusing on learning critical, relevant, and short facts essential for effective scientific claim verification.

\section{Methodology}

We introduce \modelname, a novel approach designed to enhance the performance of claim verification solutions by leveraging automatically generated short facts extracted from the evidence. We will show that \modelname is a versatile tool that can be integrated into various claim verification methods, improving the robustness and reasoning capabilities of existing models. The core of \modelname relies on weakly-labeled short facts, which are categorized as either \emph{important} for verifying a given claim or \emph{not important} for that purpose, which are used to train a multi-task learning-based model (\modelname) for importance detection and claim verification.


\subsection{Definition}

Here, we formally define the primary task of fact generation and labeling: given a claim statement $c$ and corresponding evidence statement $e$, our objective is to generate concise ``facts'' from $e$. We denote this set of facts by $\mathcal{F}_e = \{f_{1},\dots,f_m\}$. Each fact is subsequently labeled as either ``important'' or ``not important,'' denoted as $y_{f_i}\in \{\mbox{\em important}, \mbox{\em not\ important}\}$.

It is important to note that these facts are intentionally designed to be shorter in length compared to the original evidence ($e$). They serve as distilled pieces of information extracted from the broader context of the evidence. These succinct facts are intended to capture essential details or insights within the evidence, making them more manageable for claim verification tasks. An overview of \modelname is given in Figure~\ref{fig:factgen}. We next elaborate on the processes of short fact generation and weak labeling.
\subsection{Short Fact Generation}~\label{sec:factgen}
To generate short facts from the evidence $e$, we adopt a three-step approach. For these steps, we employ LLM Mistral-7B~\cite{jiang2023mistral}\footnote{Used following model checkpoint: mistralai/Mistral-7B-Instruct-v0.2}. We have experimented with different LLMs such as Vicuna-13B~\cite{vicuna2023} and GPT-3.5 and based on our experiments we observed better performance with this open-source LLM. Details of the prompts for each phase of the short fact generation using this approach are given in Appendix~\ref{sec:appendix_prompts}.

\noindent\textbf{1) Phrase matching}: Initially, we extract matching phrases from both the claim $c$ and the evidence, treating seeing each phrase as a potential answer to a questions framed around the other ($\mathcal{A} = {(a_1^c, a_1^e), \ldots, (a_n^c, a_n^e)}$). Phrases ``match'' if they convey similar meanings and/or are semantically similar. We call these answer pairs. We use an LLM to extract the matching phrases. 
We do not restrict the LLM to follow specific phrase rules such as n-grams, extracting only entities or noun phrases. This way, we ensure the capture of diverse answer pairs that are more likely to be relevant.



\noindent\textbf{2) Question Generation:} After identifying the answer pairs, we formulate concise questions from them. For each answer $a_i^c$ in the pair $(a_i^c, a_i^e)$ with corresponding claim $c$, we generate a question $q_i$. We use $c$ as the context and $a_i^c$ as a desired answer. The question does not use the evidence answer $a_i^e$ to ensure the generated question is directly associated with the claim -- because $a_i^e$ is an  answer paired with $a_i^c$, we know that the question drawn from the claim will also be aligned with the evidence answer. We create a question based on these inputs—namely, the \emph{context} and the \emph{answer} we only incorporate the answer from the claim ($a_i^c$) in this stage and not the answer from evidence ($a_i^e$). This is to 1)  ensure the generation of a high-quality question that can be associated directly with the claim, achievable only by pairing the claim with an internal answer, and 2) incorporate the essential context from the claim into the question, which will later be aligned with the $a_i^e$ for short sentence generations.


\noindent\textbf{3) Short Fact Generation}~\label{met:shortfact}: Finally, We generate short fact sentences by pairing each question $q_i$ with its corresponding evidence-based answer $a_i^e$ which was extracted in the first step and matched $a_i^c$. These questions along with the answers are then converted into full sentences $f_i$. For example, the previous question and answer results in the sentence \emph{Cellphones cause various mental health concerns for the kids.} We note that not all ($q_i$, $a_i^e$) pairs are \emph{reasonable} -- i.e., a generated $q_i$ may not align semantically well with the $a_i^e$ due to possible errors during generation or the structure of the context $c$. Therefore, to ensure a reasonable and useful fact sentence, we further refine these questions and answer pairs by querying the LLM to determine if the ($q_i$, $a_i^e$) pair is unreasonable. If the output is ``not reasonable,'' we move forward with other candidates -- i.e., ($q_{i+1}$, $a_{i+1}^e$) -- otherwise,
the sentence $f_i$ is added to the candidate answers $\mathcal{A}_c$. This step is crucial because it serves to eliminate most unsuccessful question generations that can occur with LLMs (e.g., the failures can be due to the inconsistent and hallucinated generations) and helps the \modelname to extract the most important question-answer pairs. 



\noindent\textbf{4) Weak labeling}~\label{sec:weaklabel}
Labeling each generated fact as important or not is a crucial step in the \modelname process. After extracting the candidates in the previous steps, we label a short fact sentence $f_i$ as ``important'' if the cosine similarity between $f_i$ and the claim $c$ and $f_i$ and evidence $e$ combined to exceed a predefined threshold $t$ and ``not important'' otherwise. More specifically: 
\begin{equation}
    sim(f_i,c,e) = \gamma (\cos(f_i,c)+\cos(f_i,e))
\end{equation}\label{eq:sim}
\[
y_{f_i} = \begin{dcases}
\text{``important''} & \text{if } sim(f_i,c,e) \geq t\\
\text{``not important''} & \text{otherwise}
\end{dcases}
\]

Here $\gamma$ is a hyperparameter and $\cos(.)$ is calculated using the Sentence Transformers~\cite{reimers-2019-sentence-bert} embedding of $f_i$, $c$ and $e$. 

\subsection{Joint Claim Verification and Fact Detection Framework}
Because of the success of the full context training of claim verification tasks within state-of-the-art models such as MULTIVERS~\cite{wadden-etal-2022-multivers}, PARAGRAPHJOINT~\cite{li2021paragraph}, and ARSJOINT~\cite{zhang2021abstract}, we propose a similar enhancement approach. Our framework revolves around performing full context predictions by concatenating the claim ($c$), title of the document in the scientific claim verification datasets ($t$), gold evidence ($e$), and all the facts in $\mathcal{F}_e$ with a special separator token to separate each fact in $\mathcal{F}_e$. 


The \modelname approach employs a strategy based on multitasking where the model is jointly trained to minimize a multitask loss:
\begin{equation}
L = L_{cv} + \alpha L_{fact}
\end{equation}
where $L_{cv}$ represents the cross-entropy loss associated with predicting the overall claim verification task. Specifically, we predict $y(c,e)\in\{\mbox{\em support}, \mbox{\em \ contradict}, \mbox{\em \ nei}\}$ by adding a classification head on the <$/s$> token, where $\ nei$ refers to Not Enough Info.  In addition, $L_{fact}$ denotes the binary cross-entropy loss for predicting whether each fact $f_i$ is important to the claim $c$ or not, and $\alpha$ is a hyperparameter. During inference, we only predict $y(c,e)$, setting aside the fact detection part.

\subsection{Zero-shot Claim Verification with LLMs}
In the zero-shot approach, without the need for human-annotated training dataset and finetuning a claim verification model, we leverage in-context learning ability of Large Language Models (LLMs) to extract the encoded knowledge in them using a prompting strategy aimed at eliciting the most accurate responses from them. This is done as follows. We augment \modelname generated short fact sentences $\mathcal{F_e}$ into the prompt for claim verification through fact-detection: given $c$, $e$ and $\mathcal{F}_e$ we first ask an LLM to detect the most important facts and then, by providing an explanation, we ask it to predict the verdict $y(c,e)$. 

This approach is similar to the popular Retrieval Augmented Generation \citep[RAG, see e.g.][]{lewis2020retrieval} approach used in optimizing the output of the Large Language Models using external sources. A difference between our approach to the ``retrieval'' augmented approach is that we augment the candidate facts from the evidence into the input rather than retrieving any external knowledge.   

The approach is formulated as follows: let $\mathcal{M}$ be a language model and $\mathcal{P}$ be the prompt. The $\mathcal{P}$ for the test inputs is generated by concatenating $c$, $e$ and $\mathcal{F}_e$.
We first extract \emph{important facts} and then get the predicted verdict. i.e., $p(y(c,e)|\mathcal{M}(\mathcal{P}) )$. 


\section{Experiments}
We evaluate the effect of including \modelname within different claim verification models and encoders. To evaluate this, we first explain the datasets used and introduce the baseline models we compared to our approach.  
\begin{table*}[t]
\centering
\resizebox{0.85\textwidth}{!}
{
\begin{tabular}{ @{}l l c c c |c  c c| c c c@{}}
\toprule
\multirow{2}{*}{\textbf{Setting}}&
\multirow{2}{*}{\textbf{Model}} & \multicolumn{3}{c}{\textbf{HealthVer}} &\multicolumn{3}{c}{\textbf{SciFact}} & \multicolumn{3}{c}{\textbf{SciFact-Open}}\\ 
 &  & F1 &P & R& F1 &P & R &F1 &P & R\\ [0.5ex] 
 \midrule
 \toprule
\multirow{2}{*}{\textbf{Few shot}} & 

    Longformer    & {27.8} & {25.3} &{30.7} & \underline{42.4} & \underline{43.0} &41.8 & \underline{36.2} & \underline{36.4}&36.0\\[0.5ex]
    
      &  Longformer + \modelname  &\underline{36.9}  & \underline{35.2} &\underline{38.7}&{38.3} &{35.8} &\underline{42.5}& 34.3& 28.2&\underline{43.6} \\[0.5ex]

 \midrule
\multirow{4}{*}{\textbf{Full}} 
    &Longformer   & 53.1 & 58.1 &49.1 & 54.7 &63.5 &\underline{49.0} & 40.4 & \underline{50.2} & 33.7\\[0.5ex]
    
      &  Longformer + \modelname   & \underline{53.6}&\underline{58.2}& \underline{49.6}&
      
      \underline{56.3}&\underline{67.2} &48.5&
      
      \underline{43.1}& {49.7}& \underline{38.1}\\[0.5ex]
      \cline{2-11}
\cline{2-6}

     &MULTIVERS  & 60.6& {59.1} &  \textbf{62.0}& 70.4& \textbf{70.8}&70.0 &
     \textbf{65.0} &\textbf{65.3}&\textbf{64.8} \\[0.5ex]
     
     &MULTIVERS + \modelname  & \textbf{61.2} & \textbf{64.5} &{58.2} &{70.4} & 70.3& \textbf{70.3}&
     {61.1} &{62.6}&59.7\\[0.5ex]
 \hline
\end{tabular}
}
\caption{\label{res:main_finetune}\small{ Overall performance comparison between different baselines without and with (+\modelname) multi-task learning incorporating \modelname. SciFact-Open results are reported in a zero-shot setting. The best results for each dataset are highlighted in bold and the best results within each pair (with and without \modelname) are underlined.} }
\end{table*}

\subsection{Datasets}

\noindent \textbf{SciFact}~\cite{wadden-etal-2020-fact} consists of expert annotated scientific claims from biomedical literature with corresponding evidence sentences retrieved from abstracts. \emph{Supported} claims are human-generated using abstract citation sentences, and \emph{Contradicted} claims negate original claims. 

\noindent \textbf{SciFact-Open}~\cite{wadden-etal-2022-scifact}  constitutes a test collection specifically crafted for the assessment of scientific claim verification systems. In addition to the task of verifying claims against evidence within the SciFact domain, this dataset contains evidence originating from a vast scientific corpus of 500,000 documents. 

\noindent \textbf{HealthVer}~\cite{sarrouti2021evidence} is a compilation of COVID-19-related claims from real-world scenarios that have been subjected to fact-checking using scientific articles. Unlike most available datasets, where \emph{contradict}ed claims are usually just the negation of the supported ones, in this dataset \emph{contradicted} claims are themselves extracted from real-world claims.  The claims in this dataset are more challenging compared to other datasets. 
More detailed statistics of the datasets are given in Appendix~\ref{a:datastats}. 






\subsection{Baselines}

We evaluate \modelname in supervised and zero-shot settings. In a supervised setting, we either fully or \emph{few-shot} train the state-of-the-art models on the given datasets. For the zero-shot setting, we use several best-performing LLMs and prompt them to predict the verdict based on different baseline prompting strategies. For few-shot supervised training, we train on $k=45$ training samples.

\subsubsection{Supervised Baselines}

We incorporate \modelname as an add-on for a multi-task learning-based approach on two transformer-based encoders.
We train the supervised models on NVIDIA RTX8000 GPU and overall model parameters do not exceed 1B. We set the learning rate to $2e-5$ and save the best model in $25$ epochs. We choose $0.5$ for the $\gamma$ similarity parameter, in equation (1) and $10$ \footnote{We performed experiments with $5$, $10$ and $15$ and the best performing value was $15$.} for the $\alpha$ hyperparameter of equation (2). The threshold $t$ for the cosine similarity between fact sentences and claim and evidence is set to $0.6$. 


\noindent \textbf{Longformer}~\cite{Beltagy2020Longformer} With the self-attention mechanism incorporated into this model and its ability to process long sequences, we use this encoder to concatenate short sentences into the claim along with additional context provided in the title (if any).

\noindent \textbf{MULTIVERS} ~\cite{wadden-etal-2022-multivers} is a state-of-the-art supervised scientific claim verification approach which uses Longformer as a base encoder for long-context end-to-end claim verification in a multi-task learning based approach where in addition to the claim and title it incorporates the whole document (abstract) for both claim verification and rationale (evidence) selection. We augment the short sentences extracted by \modelname into the model as an input and train \modelname on top of MULTIVERS in a multitasking-based approach.

\subsubsection{Zero-shot baselines}
LLMs serve as a robust source of knowledge and demonstrate impressive outcomes in various downstream tasks, especially in contexts where zero-shot and few-shot learning are employed. However, the effectiveness of these models heavily depends on the methods used to prompt their responses. Consequently, we evaluate state-of-the-art prompting methods both specific to the claim verification task and general task approaches, and compare them to our novel prompting method based on adding the \modelname-generated short sentences into the prompt and requiring the LLM to detect the most important sentences for verdict as well as predicting the verdict. We name this prompting strategy Aug\modelname. More details of this strategy are given in Appendix~\ref{a_aug}. Below are the baseline prompting strategies used to compare with Aug\modelname in the experiments.

\noindent \textbf{Vanilla}: We engage LLMs to assess the truthfulness of claims based on provided evidence and to offer justifications for their verdicts. This process is carried out without integrating any extra knowledge or employing a specific strategy.

\noindent \textbf{Chain of Thought (CoT)}~\cite{wei2022chain} This popular approach involves breaking down the task into a series of logical steps presented to LLMs via prompts for the given context. We use this approach by providing the claim and evidence as input and instructing it to think step by step and provide an explanation before predicting the verdict. We consequently add the \textit{let's think step by step} instruction into the prompt and provide a few shot examples where the verdict is given followed by a step-by-step reasoning explanations. 
We compare these baseline strategies in FlanT5-XXL~\cite{chung2022scaling}, GPT-3.5 (gpt-3.5-turbo checkpoint),, Llama2-13B (Llama-2-13b-chat-hf checkpoint) ~\cite{touvron2023llama}, Vicuna-13B~\cite{vicuna2023} (vicuna-13b-v1.5 checkpoint), and Mistral-7B Instruct (Mistral-7B-Instruct-v0.2 checkpoint). We perform experiments in few-shot prompting ($k=5$) for all the strategies. Details of the prompts for Vanilla and CoT are given in Appendix ~\ref{a:prompts}. 

\noindent \textbf{ProgramFC} ~\cite{pan-etal-2023-fact} is a newly introduced approach that converts complex claims into sub-claims which are then used to generate reasoning programs using LLMs that are executed and used for guiding the verification. We utilize the closed-book setting of this method with N=1.  This approach is built for only two-label datasets where claims are either \emph{supported} or \emph{contradicted} by evidence. We used GPT-3.5 to generate programs for ProgramFC and extracted the verification with FlanT5-XL. We experimented with this model in two-label settings (\emph{supported} and \emph{contradicted}) because the original model is designed in binary verification mode. For a fair comparison, we report binary classification results (by excluding the \emph{not enough info} labeled dataset) in all our experiments as well.

 \begin{table*}[ht]
\centering

\resizebox{0.9\textwidth}{!}{
\begin{tabular}{@{}llcccccc@{}}
\toprule
\multicolumn{1}{c}{Datasets} & &
  \multicolumn{2}{c}{SciFact} &
  \multicolumn{2}{c}{SciFact-Open} & \multicolumn{2}{c}{HealthVer} \\ 
 
  \cmidrule(r){3-4} \cmidrule(l){5-6} \cmidrule(l){7-8} 
\multicolumn{1}{c}{Metrics} & &
   F1 &
  F1 /wo NEI &
  F1 &
  F1 /wo NEI &
F1 &
  F1 /wo NEI \\ \midrule
\multirow{3}{*}{FlanT5-XXL$^*$}  & Vanilla &
  \underline{75.4} &\textbf{84.4$^*$}  &
  68.5 & \underline{84.3} &
  50.5 & 69.1 \\
  & CoT &
  67.9 & 82.6 &
  68.5 &  83.2 &
  53.6 &62.4  \\
  & Aug\modelname &
  74.5 & 82.4 &
  \underline{73.6} & 83.4 &
\underline{56.5} & \underline{69.1} \\

  \cmidrule(r){2-8}
\multirow{3}{*}{Llama2-13B$^*$}  & Vanilla &
  47.7 & 63.1 &
  47.4 &61.0 &
  48.9 & 67.3 \\
  & CoT &
  55.4 & 65.7 &
  55.1 & 71.5 &
  51.5 & 65.5 \\
  & Aug\modelname &
 \underline{75.1} &\underline{71.7} &
  \underline{70.5} &\underline{76.7} &
 \textbf{62.3$^*$}  & \textbf{75.8$^*$} \\

\cmidrule(r){2-8}
  \multirow{3}{*}{Vicuna-13B$^*$}  & Vanilla &
  38.4 & 67.2 &
  \underline{53.5} &68.2 &
  51.0 &58.7  \\
  & CoT &
  45.3 & 61.5 &
  52.7 &70.9 &
  50.4 & 62.0\\
  & Aug\modelname &
 \underline{49.1} &\underline{75.8} &
  50.3 &\underline{79.5} &
 \underline{51.3} & \underline{71.8} \\

  \cmidrule(r){2-8}

    \multirow{3}{*}{Mistral-7B$^*$}  & Vanilla &
 67.3 &79.0 &
 62.5  & 81.8 &
 51.0 &73.0  \\
  & CoT &
  70.8 &80.3 &
  65.0 &\underline{83.3} &
  54.2 &\underline{73.8} \\
  & Aug\modelname &
  \textbf{76.0$^*$} & \underline{82.3}
  & 
  \textbf{76.0$^*$} &{82.4}
   &
 \underline{ 61.8} & 73.6 \\

 \cmidrule(r){2-8}
  \multirow{3}{*}{GPT-3.5}  & Vanilla &
  64.5 &72.5&
  63.0 &80.4 &
  50.9 &\underline{68.0} \\
  & CoT &
  69.8 &\underline{81.8} &
  62.9 &\textbf{84.5$^*$} &
  52.1 &67.9 \\
  & Aug\modelname &
  \underline{75.4} & 70.2 &
  \underline{71.6} & 73.1&
  \underline{58.6} & 64.9 \\



\bottomrule
  ProgramFC&  &
  $-$ &45.0 &
  $-$ & 78.0 &
 $-$ &62.9 \\
 \bottomrule
\end{tabular}
}
\caption{\small{We evaluate the effectiveness of different prompting strategies in 5 LLMs. We report results both with \emph{not enough info} data samples and without them (/wo NEI). For open source LLMs, we ran experiments 5 times and report the average scores (indicated with $*$). The best-performing strategy for each LLM is underlined and overall the best results are highlighted in bold for each dataset. Statistically significant ($p<0.05$) results compared to the best-performing ones are highlighted with $^*$.}}\label{res:prompts}
\end{table*}

\subsection{Main Results}

\subsubsection{Supervised Setup} We first report the results of \emph{supervised} baselines with and without \modelname incorporated in their training process in Table~\ref{res:main_finetune}. We experiment with few-shot and full training setups. We observe that incorporating \modelname into the Longformer encoder achieves the best performance in all three datasets (in bold) in the Full training setup. The average performance gain in F1 when adding \modelname to Longformer is 3.0$\%$  for SciFact. Longformer + \modelname in the few-shot setting also improves the F1 score for HealthVer by  32.7$\%$. However, we do not see a performance improvement in the few-shot setting for SciFact and SciFact-Open datasets. 
As mentioned earlier, the results of SciFact-Open dataset are reported in a zero-shot setting (with model trained on SciFact training dataset), resulting in lower performance. Additionally, SciFact-Open receives less benefit from \modelname than other datasets even in the cases where it does improve results. We suspect that this is due to the more complex nature of the dataset, because it contains claims that are both \emph{supported} and \emph{contradicted} by different evidence sentences. The outcomes are consistent with the top-performing baseline, MULTIVERS. By integrating \modelname into MULTIVERS, we achieve similar performance, despite the advantage of complete context encoding within this framework.

\subsubsection{Zero-shot Setup} 

 The results  corresponding to the performance evaluation for the zero-shot prompting with different strategies  
are reported in Table~\ref{res:prompts}. 

We observe that Aug\modelname significantly improves the performance of Llama2-13B, Mistral-7B, and GPT-3.5 in all three datasets compared to the best-performing baseline with an average performance gain of $28.1\%$, $12.7\%$ and $11.3\%$ in the F1 score for SciFact, Scifact-Open, and Healthver test sets respectively. Similarly, Aug\modelname shows significant improvements for Vicuna-13B in SciFact and HealthVer and FlanT5-XXL with Aug\modelname outperforms other prompting strategies in Scifact-Open and HealthVer test sets. Comparison between ProgramFC and baselines also shows the limited advantage in predicting verdicts in scientific claim verification datasets compared to the general claim verification datasets.

Overall Aug\modelname demonstrates better performance compared to other prompting strategies which suggests the effectiveness of the short fact generation strategy based on the connection between claim and evidence and its performance is comparable to the best-performing baseline in the binary setting.

\begin{figure*}[ht]
    \centering
    \includegraphics[width=\textwidth]{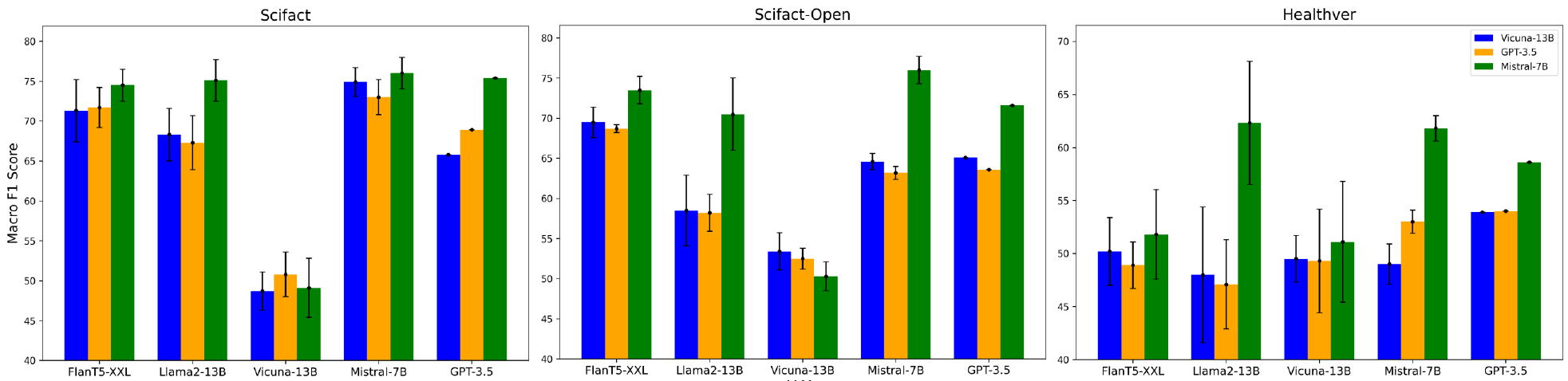}
    \caption{\small{Comparing the F1 Score of zero-shot claim verification task on three test sets when \modelname is generated with three different LLMs (Vicuna-13B, GPT-3.5 and Mistral-7B).   }}
    \label{fig:assess}
\end{figure*}

\begin{figure}[ht]
    \centering
    \includegraphics[width=0.9\columnwidth]{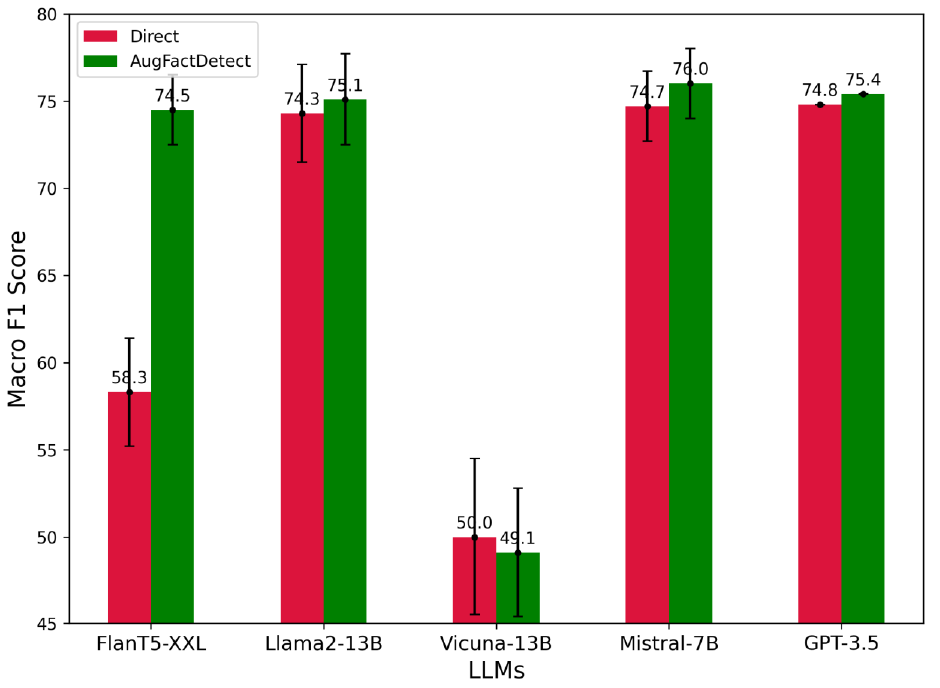}
    \caption{\small{Comparison in Macro F1 score for SciFact between Aug\modelname and Direct.}}
    \label{fig:directvsfactdetect}
\end{figure}
\subsection{Effectiveness of \modelname} 
To further understand the impact of the \modelname, we compare \modelname based short fact generation approach with the Direct approach where we directly generate short sentences from evidence $e$ (we give 5 examples as few-shot prompting). The details of the promoting strategy and the examples are given in Appendix~\ref{app:direct}.  We collect the short sentences for each piece of evidence in a claim-evidence (CE) pair, for the SciFact dataset (dev set) and run experiments in the zero-shot setup for 5 LLMS. Macro F1 score comparisons between Direct and Aug\modelname are given in Figure~\ref{fig:directvsfactdetect}. We report results in an average of 5 runs.

Overall, Aug\modelname performs better compared to the Direct approach across 4 out of 5 LLMs with a significant difference in FlanT5-XXL and Mistral-7B. These results suggest the usefulness of the three-step approach compared to the baseline direct sentence generation approach. We hypothesize that one key reason for this is in the Direct approach, the generated sentences are based on the evidence only without making a meaningful connection between the claim and the evidence. Therefore, effective short sentences based on the keyphrases linking claim and evidence provide an advantage in predicting the verdict.

\subsection{Assessing Generation Quality for \modelname} 
Here, we explore the impact of various underlying large language models (LLMs) on the quality of \modelname generated short sentences. We evaluate this by regenerating short fact sentences using three different LLMs: Mistral-7B\footnote{checkpoint: Mistral-7B-Instruct-v0.2}, GPT-3.5\footnote{{checkpoint: gpt-3.5-turbo-1106}}, and Vicuna-13B \footnote{checkpoint: vicuna-13b-v1.5} and assess their effect in the performance of Aug\modelname for the claim verification task. The findings are depicted in Figure~\ref{fig:assess}.

The results indicate that choosing Vicuna-13B and GPT-3.5 as the base models for short fact generation demonstrates approximately similar performance across 5 LLMs for all the test sets whereas, Mistral-7B exhibits more pronounced performance. Even though Mistral-7B is a relatively smaller model, shows sufficient and consistent performance gains for the claim verification task whereas, the performance drops with using Vicuna-13B and GPT-3.5 as base models for short fact-generation. This result is independent of the LLM parameter and quality and based on our manual analysis we observed that GPT-3.5 and Vicuna-13B show higher sensitivity to the ``reasonability filter'' and many question-answer pairs generated in the question generation phase (see ~\ref{met:shortfact}) are marked as not reasonable and do not make it to the next phase of sentence generation resulting in an average low number of generated sentences compared to generated sentences using Mistral-7B with 0.47 and 2.31 for GPT-3.5 and Vicuna-13B compared to 3.64 average number of short sentences per CE pair for Mistral-7B. We additionally perform a human analysis for the overall quality of generated sentences which we detail in Appendix~\ref{a:human}.


\section{Conclusion and Future Work}
In this work, we propose \modelname, an effective short fact generation technique, for comprehensive and high-quality condensed small sentences derived from evidence.  With the relevance-based weak-labeling approach this dataset can be augmented to any state-of-the-art claim verification model as a multi-task learning to train fact detection and claim verification. The effectiveness of this model has been demonstrated in both fine-tuned and prompt-based models. Our results suggest that \modelname incorporated claim-verification task in a zero-shot setting consistently improves performance on average by $17.3\%$ across three challenging scientific claim verification test sets.  

\modelname can have broader applications in different fact-checking and factual consistency evaluation tasks. As a future work, we plan to incorporate \modelname in the factual consistency evaluation of LLMs. Our preliminary results (see Appendix~\ref{a:factuality}) showed promising performance for factuality evaluation in FIB~\cite{tam2022fib} dataset.  

\section{Limitations}
A drawback of our method is the reliance on a generative language model for producing short fact sentences throughout the entire process. Despite employing Mistral-7B, which is among the top open-source LLMs available, the factual accuracy and overall quality of the generated content are bounded by the capabilities of this particular model. Consequently, any inaccuracies from the model could impact the effectiveness of the end-to-end claim verification system. 

Furthermore, a limitation of zero-shot FactDetect in real-world claim-verification systems is the need to augment the short sentences into the prompt, which is an additional step and can be time-consuming in the claim verification task. However, this problem is mitigated when we fine-tune a claim-verification system with \modelname in the training phase, and during inference, we just use the claim and evidence as input.

\section{Ethics Statement
}

\noindent \textbf{Biases.} We acknowledge the possibility of bias in generated outputs from the trained LLM. However, this is beyond our control.

\noindent \textbf{Potential Risks.} Our approach can be used for automated fact-checking. However, they could also be used by malicious actors to manipulate and attack fact-checking models. A possible future direction is to detect such malicious actions before deployment.

\noindent\textbf{Environmental Impact.} Training and using LLMs involves considerable computational resources, including the necessity for GPUs or TPUs during training or inference which can have an impact on the environment. However, we trained our datasets on relatively smaller language models with less than 1B parameters and we used LLMs for inference only which has negligible negative effect on the environment.

\bibliography{custom}

\begin{thebibliography}{34}
\expandafter\ifx\csname natexlab\endcsname\relax\def\natexlab#1{#1}\fi

\bibitem[{Beltagy et~al.(2020)Beltagy, Peters, and Cohan}]{Beltagy2020Longformer}
Iz~Beltagy, Matthew~E. Peters, and Arman Cohan. 2020.
\newblock Longformer: The long-document transformer.
\newblock \emph{arXiv:2004.05150}.

\bibitem[{Chiang et~al.(2023)Chiang, Li, Lin, Sheng, Wu, Zhang, Zheng, Zhuang, Zhuang, Gonzalez, Stoica, and Xing}]{vicuna2023}
Wei-Lin Chiang, Zhuohan Li, Zi~Lin, Ying Sheng, Zhanghao Wu, Hao Zhang, Lianmin Zheng, Siyuan Zhuang, Yonghao Zhuang, Joseph~E. Gonzalez, Ion Stoica, and Eric~P. Xing. 2023.
\newblock \href {https://lmsys.org/blog/2023-03-30-vicuna/} {Vicuna: An open-source chatbot impressing gpt-4 with 90\%* chatgpt quality}.

\bibitem[{Chung et~al.(2022)Chung, Hou, Longpre, Zoph, Tay, Fedus, Li, Wang, Dehghani, Brahma et~al.}]{chung2022scaling}
Hyung~Won Chung, Le~Hou, Shayne Longpre, Barret Zoph, Yi~Tay, William Fedus, Yunxuan Li, Xuezhi Wang, Mostafa Dehghani, Siddhartha Brahma, et~al. 2022.
\newblock Scaling instruction-finetuned language models.
\newblock \emph{arXiv preprint arXiv:2210.11416}.

\bibitem[{Cui et~al.(2019)Cui, Shu, Wang, Lee, and Liu}]{10.1145/3357384.3357862}
Limeng Cui, Kai Shu, Suhang Wang, Dongwon Lee, and Huan Liu. 2019.
\newblock \href {https://doi.org/10.1145/3357384.3357862} {Defend: A system for explainable fake news detection}.
\newblock In \emph{Proceedings of the 28th ACM International Conference on Information and Knowledge Management}, CIKM '19, page 2961–2964, New York, NY, USA. Association for Computing Machinery.

\bibitem[{Dai et~al.(2022)Dai, Hsu, Xiong, and Ku}]{dai2022ask}
Shih-Chieh Dai, Yi-Li Hsu, Aiping Xiong, and Lun-Wei Ku. 2022.
\newblock Ask to know more: Generating counterfactual explanations for fake claims.
\newblock In \emph{Proceedings of the 28th ACM SIGKDD Conference on Knowledge Discovery and Data Mining}, pages 2800--2810.

\bibitem[{Diggelmann et~al.(2020)Diggelmann, Boyd-Graber, Bulian, Ciaramita, and Leippold}]{diggelmann2020climatefever}
Thomas Diggelmann, Jordan Boyd-Graber, Jannis Bulian, Massimiliano Ciaramita, and Markus Leippold. 2020.
\newblock \href {http://arxiv.org/abs/2012.00614} {Climate-fever: A dataset for verification of real-world climate claims}.

\bibitem[{Guo et~al.(2022)Guo, Schlichtkrull, and Vlachos}]{guo2022survey}
Zhijiang Guo, Michael Schlichtkrull, and Andreas Vlachos. 2022.
\newblock A survey on automated fact-checking.
\newblock \emph{Transactions of the Association for Computational Linguistics}, 10:178--206.

\bibitem[{Hermann et~al.(2015)Hermann, Kocisky, Grefenstette, Espeholt, Kay, Suleyman, and Blunsom}]{hermann2015teaching}
Karl~Moritz Hermann, Tomas Kocisky, Edward Grefenstette, Lasse Espeholt, Will Kay, Mustafa Suleyman, and Phil Blunsom. 2015.
\newblock Teaching machines to read and comprehend.
\newblock \emph{Advances in neural information processing systems}, 28.

\bibitem[{Jiang et~al.(2023)Jiang, Sablayrolles, Mensch, Bamford, Chaplot, Casas, Bressand, Lengyel, Lample, Saulnier et~al.}]{jiang2023mistral}
Albert~Q Jiang, Alexandre Sablayrolles, Arthur Mensch, Chris Bamford, Devendra~Singh Chaplot, Diego de~las Casas, Florian Bressand, Gianna Lengyel, Guillaume Lample, Lucile Saulnier, et~al. 2023.
\newblock Mistral 7b.
\newblock \emph{arXiv preprint arXiv:2310.06825}.

\bibitem[{Jolly et~al.(2022)Jolly, Atanasova, and Augenstein}]{info13100500}
Shailza Jolly, Pepa Atanasova, and Isabelle Augenstein. 2022.
\newblock \href {https://doi.org/10.3390/info13100500} {Generating fluent fact checking explanations with unsupervised post-editing}.
\newblock \emph{Information}, 13(10).

\bibitem[{Kotonya and Toni(2020)}]{kotonya-toni-2020-explainable-automated}
Neema Kotonya and Francesca Toni. 2020.
\newblock \href {https://doi.org/10.18653/v1/2020.emnlp-main.623} {Explainable automated fact-checking for public health claims}.
\newblock In \emph{Proceedings of the 2020 Conference on Empirical Methods in Natural Language Processing (EMNLP)}, pages 7740--7754, Online. Association for Computational Linguistics.

\bibitem[{Krishna et~al.(2022)Krishna, Riedel, and Vlachos}]{krishna2022proofver}
Amrith Krishna, Sebastian Riedel, and Andreas Vlachos. 2022.
\newblock Proofver: Natural logic theorem proving for fact verification.
\newblock \emph{Transactions of the Association for Computational Linguistics}, 10:1013--1030.

\bibitem[{Lee et~al.(2021)Lee, Won, Kim, Lee, Park, and Jung}]{lee2021crossaug}
Minwoo Lee, Seungpil Won, Juae Kim, Hwanhee Lee, Cheoneum Park, and Kyomin Jung. 2021.
\newblock Crossaug: A contrastive data augmentation method for debiasing fact verification models.
\newblock In \emph{Proceedings of the 30th ACM International Conference on Information \& Knowledge Management}, CIKM '21. Association for Computing Machinery.

\bibitem[{Lewis et~al.(2020)Lewis, Perez, Piktus, Petroni, Karpukhin, Goyal, K{\"u}ttler, Lewis, Yih, Rockt{\"a}schel et~al.}]{lewis2020retrieval}
Patrick Lewis, Ethan Perez, Aleksandra Piktus, Fabio Petroni, Vladimir Karpukhin, Naman Goyal, Heinrich K{\"u}ttler, Mike Lewis, Wen-tau Yih, Tim Rockt{\"a}schel, et~al. 2020.
\newblock Retrieval-augmented generation for knowledge-intensive nlp tasks.
\newblock \emph{Advances in Neural Information Processing Systems}, 33:9459--9474.

\bibitem[{Li et~al.(2021)Li, Burns, and Peng}]{li2021paragraph}
Xiangci Li, Gully~A Burns, and Nanyun Peng. 2021.
\newblock A paragraph-level multi-task learning model for scientific fact-verification.
\newblock In \emph{SDU@ AAAI}.

\bibitem[{Mehta et~al.(2022)Mehta, Rangwala, and Ramakrishnan}]{mehta2022improving}
Sneha Mehta, Huzefa Rangwala, and Naren Ramakrishnan. 2022.
\newblock Improving zero-shot event extraction via sentence simplification.
\newblock \emph{arXiv preprint arXiv:2204.02531}.

\bibitem[{Narayan et~al.(2018)Narayan, Cohen, and Lapata}]{narayan2018don}
Shashi Narayan, Shay~B Cohen, and Mirella Lapata. 2018.
\newblock Don't give me the details, just the summary! topic-aware convolutional neural networks for extreme summarization.
\newblock \emph{arXiv preprint arXiv:1808.08745}.

\bibitem[{Pan et~al.(2021)Pan, Chen, Xiong, Kan, and Wang}]{pan-etal-2021-Zero-shot-FV}
Liangming Pan, Wenhu Chen, Wenhan Xiong, Min-Yen Kan, and William~Yang Wang. 2021.
\newblock Zero-shot fact verification by claim generation.
\newblock In \emph{The Joint Conference of the 59th Annual Meeting of the Association for Computational Linguistics and the 11th International Joint Conference on Natural Language Processing (ACL-IJCNLP 2021)}, Online.

\bibitem[{Pan et~al.(2023)Pan, Wu, Lu, Luu, Wang, Kan, and Nakov}]{pan-etal-2023-fact}
Liangming Pan, Xiaobao Wu, Xinyuan Lu, Anh~Tuan Luu, William~Yang Wang, Min-Yen Kan, and Preslav Nakov. 2023.
\newblock \href {https://doi.org/10.18653/v1/2023.acl-long.386} {Fact-checking complex claims with program-guided reasoning}.
\newblock In \emph{Proceedings of the 61st Annual Meeting of the Association for Computational Linguistics (Volume 1: Long Papers)}, pages 6981--7004, Toronto, Canada. Association for Computational Linguistics.

\bibitem[{Popat et~al.(2017)Popat, Mukherjee, Str{\"o}tgen, and Weikum}]{popat2017truth}
Kashyap Popat, Subhabrata Mukherjee, Jannik Str{\"o}tgen, and Gerhard Weikum. 2017.
\newblock Where the truth lies: Explaining the credibility of emerging claims on the web and social media.
\newblock In \emph{Proceedings of the 26th International Conference on World Wide Web Companion}, pages 1003--1012.

\bibitem[{Reimers and Gurevych(2019)}]{reimers-2019-sentence-bert}
Nils Reimers and Iryna Gurevych. 2019.
\newblock \href {https://arxiv.org/abs/1908.10084} {Sentence-bert: Sentence embeddings using siamese bert-networks}.
\newblock In \emph{Proceedings of the 2019 Conference on Empirical Methods in Natural Language Processing}. Association for Computational Linguistics.

\bibitem[{Saakyan et~al.(2021)Saakyan, Chakrabarty, and Muresan}]{saakyan2021covid}
Arkadiy Saakyan, Tuhin Chakrabarty, and Smaranda Muresan. 2021.
\newblock Covid-fact: Fact extraction and verification of real-world claims on covid-19 pandemic.
\newblock \emph{arXiv preprint arXiv:2106.03794}.

\bibitem[{Sarrouti et~al.(2021)Sarrouti, Abacha, M’rabet, and Demner-Fushman}]{sarrouti2021evidence}
Mourad Sarrouti, Asma~Ben Abacha, Yassine M’rabet, and Dina Demner-Fushman. 2021.
\newblock Evidence-based fact-checking of health-related claims.
\newblock In \emph{Findings of the Association for Computational Linguistics: EMNLP 2021}, pages 3499--3512.

\bibitem[{Stammbach and Ash(2020)}]{stammbach2020fever}
Dominik Stammbach and Elliott Ash. 2020.
\newblock e-fever: Explanations and summaries for automated fact checking.
\newblock \emph{Proceedings of the 2020 Truth and Trust Online (TTO 2020)}, pages 32--43.

\bibitem[{Tam et~al.(2022)Tam, Mascarenhas, Zhang, Kwan, Bansal, and Raffel}]{tam2022fib}
Derek Tam, Anisha Mascarenhas, Shiyue Zhang, Sarah Kwan, Mohit Bansal, and Colin Raffel. 2022.
\newblock Evaluating the factual consistency of large language models through summarization.
\newblock \emph{arXiv preprint arXiv:2211.08412}.

\bibitem[{Thorne et~al.(2018)Thorne, Vlachos, Christodoulopoulos, and Mittal}]{Thorne18Fever}
James Thorne, Andreas Vlachos, Christos Christodoulopoulos, and Arpit Mittal. 2018.
\newblock {FEVER}: a large-scale dataset for fact extraction and {VERification}.
\newblock In \emph{NAACL-HLT}.

\bibitem[{Touvron et~al.(2023)Touvron, Martin, Stone, Albert, Almahairi, Babaei, Bashlykov, Batra, Bhargava, Bhosale et~al.}]{touvron2023llama}
Hugo Touvron, Louis Martin, Kevin Stone, Peter Albert, Amjad Almahairi, Yasmine Babaei, Nikolay Bashlykov, Soumya Batra, Prajjwal Bhargava, Shruti Bhosale, et~al. 2023.
\newblock Llama 2: Open foundation and fine-tuned chat models.
\newblock \emph{arXiv preprint arXiv:2307.09288}.

\bibitem[{Wadden et~al.(2020)Wadden, Lin, Lo, Wang, van Zuylen, Cohan, and Hajishirzi}]{wadden-etal-2020-fact}
David Wadden, Shanchuan Lin, Kyle Lo, Lucy~Lu Wang, Madeleine van Zuylen, Arman Cohan, and Hannaneh Hajishirzi. 2020.
\newblock \href {https://doi.org/10.18653/v1/2020.emnlp-main.609} {Fact or fiction: Verifying scientific claims}.
\newblock In \emph{Proceedings of the 2020 Conference on Empirical Methods in Natural Language Processing (EMNLP)}, pages 7534--7550, Online. Association for Computational Linguistics.

\bibitem[{Wadden et~al.(2022{\natexlab{a}})Wadden, Lo, Kuehl, Cohan, Beltagy, Wang, and Hajishirzi}]{wadden-etal-2022-scifact}
David Wadden, Kyle Lo, Bailey Kuehl, Arman Cohan, Iz~Beltagy, Lucy~Lu Wang, and Hannaneh Hajishirzi. 2022{\natexlab{a}}.
\newblock \href {https://doi.org/10.18653/v1/2022.findings-emnlp.347} {{S}ci{F}act-open: Towards open-domain scientific claim verification}.
\newblock In \emph{Findings of the Association for Computational Linguistics: EMNLP 2022}, pages 4719--4734, Abu Dhabi, United Arab Emirates. Association for Computational Linguistics.

\bibitem[{Wadden et~al.(2022{\natexlab{b}})Wadden, Lo, Wang, Cohan, Beltagy, and Hajishirzi}]{wadden-etal-2022-multivers}
David Wadden, Kyle Lo, Lucy~Lu Wang, Arman Cohan, Iz~Beltagy, and Hannaneh Hajishirzi. 2022{\natexlab{b}}.
\newblock \href {https://doi.org/10.18653/v1/2022.findings-naacl.6} {{M}ulti{V}er{S}: Improving scientific claim verification with weak supervision and full-document context}.
\newblock In \emph{Findings of the Association for Computational Linguistics: NAACL 2022}, pages 61--76, Seattle, United States. Association for Computational Linguistics.

\bibitem[{Wang and Shu(2023)}]{wang2023explainable}
Haoran Wang and Kai Shu. 2023.
\newblock Explainable claim verification via knowledge-grounded reasoning with large language models.
\newblock \emph{arXiv preprint arXiv:2310.05253}.

\bibitem[{Wei et~al.(2022)Wei, Wang, Schuurmans, Bosma, Xia, Chi, Le, Zhou et~al.}]{wei2022chain}
Jason Wei, Xuezhi Wang, Dale Schuurmans, Maarten Bosma, Fei Xia, Ed~Chi, Quoc~V Le, Denny Zhou, et~al. 2022.
\newblock Chain-of-thought prompting elicits reasoning in large language models.
\newblock \emph{Advances in Neural Information Processing Systems}, 35:24824--24837.

\bibitem[{Yang et~al.(2019)Yang, Pentyala, Mohseni, Du, Yuan, Linder, Ragan, Ji, and Hu}]{10.1145/3308558.3314119}
Fan Yang, Shiva~K. Pentyala, Sina Mohseni, Mengnan Du, Hao Yuan, Rhema Linder, Eric~D. Ragan, Shuiwang Ji, and Xia~(Ben) Hu. 2019.
\newblock \href {https://doi.org/10.1145/3308558.3314119} {Xfake: Explainable fake news detector with visualizations}.
\newblock In \emph{The World Wide Web Conference}, WWW '19, page 3600–3604, New York, NY, USA. Association for Computing Machinery.

\bibitem[{Zhang et~al.(2021)Zhang, Li, Fukumoto, and Ye}]{zhang2021abstract}
Zhiwei Zhang, Jiyi Li, Fumiyo Fukumoto, and Yanming Ye. 2021.
\newblock Abstract, rationale, stance: a joint model for scientific claim verification.
\newblock \emph{arXiv preprint arXiv:2110.15116}.

\end{thebibliography}
\bibliographystyle{acl_natbib}

\appendix
\begin{table*}[ht]
\centering
\resizebox{0.75\textwidth}{!}{
\begin{tabular}{l l cc cc cc}
\toprule
& &\multicolumn{2}{c}{Train} & \multicolumn{2}{c}{Dev}  & \multicolumn{2}{c}{Test}\\
\cline{3-4} \cline{5-6} \cline{7-8}
Dataset& Corpus & Claims & CE pairs & Claims & CE pairs & Claims  & CE pairs \\
\midrule
SciFact-Open & 500K & $-$ & $-$ &$-$& $-$ & 279 & 460 \\
Scifact & 14K & 809 & 564 & 300 & 209 & 300 & $-$  \\
HealthVer & 322 & 1393 &3340 &230 & 508 & 230& 599\\
\bottomrule
\end{tabular}
}
\caption{\small{Statistics of datasets used in our experiments. Claim Evidence pairs (CE pairs) for each dataset are provided. Scifact test set is not included with gold-labeled evidence sentences therefore the CE pairs are not reported for this dataset. }}
\label{data:stats}
\end{table*}
\section{Details in Short Fact Generation}
\label{sec:appendix_prompts}
\subsection{Prompt for Matching Key Phrase Extraction}

\begin{figure}[ht]
    \centering
    \includegraphics[width=0.9\columnwidth]{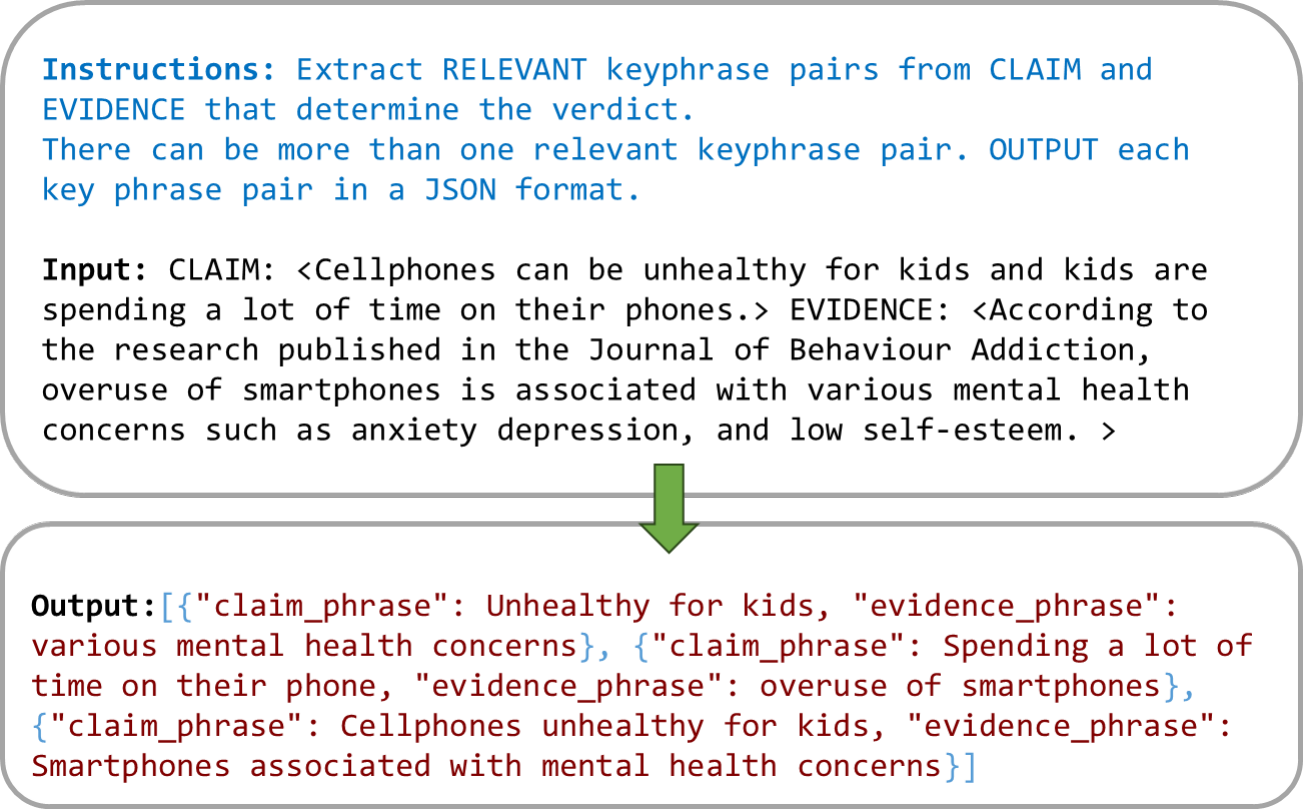}
    \caption{Example of the prompting method used to extract matching key phrases between claim $c$ and evidence $e$.}
    \label{fig:cestars}
\end{figure}

Figure~\ref{fig:cestars} provides an example of a prompt used for key-phrase extraction. 

\subsection{Prompt Strategy for Question Generation}
\begin{figure}[ht]
    \centering
    \includegraphics[width=0.9\columnwidth]{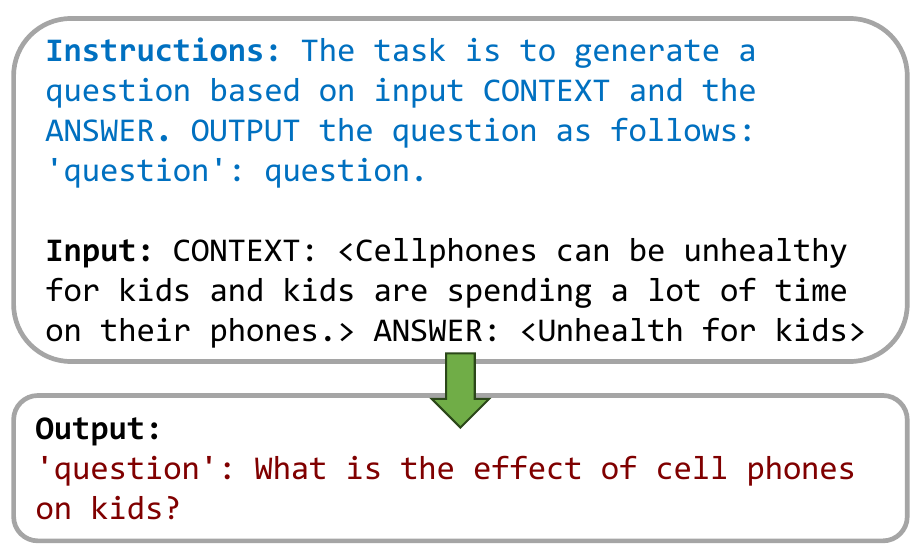}
    \caption{Example of the prompting method used to extract question from a claim $c$ as context and $a^c_i$ as answer.}
    \label{fig:q_prompt}
\end{figure}
Figure~\ref{fig:q_prompt} provides an example of the prompt strategy used to generate a question from extracted phrases from claim and an answer extracted from the previous step. We use a standard question generation prompting method in this step. 

\subsection{Prompt for Short Fact Generation from Question and Answer}
\begin{figure}[ht]
    \centering
    \includegraphics[width=0.9\columnwidth]{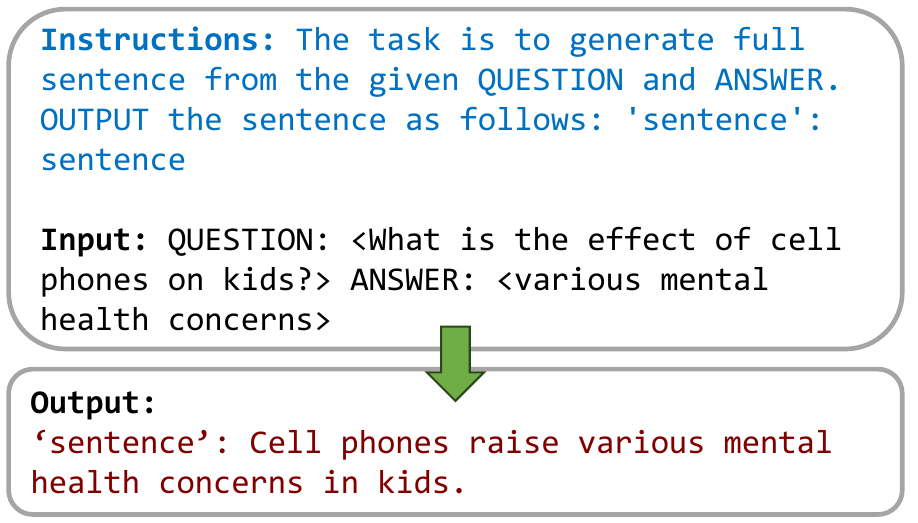}
    \caption{Example of the prompting method used to extract short sentence from a question $q_i$ and $a^e_i$.}
    \label{fig:f_prompt}
\end{figure}
Figure ~\ref{fig:f_prompt} provides an example of the prompting method used to extract the short sentence, final step in short fact generation, from the generated question and matching evidence phrase.

\section{Dataset statistics}\label{a:datastats}
Statistics of the scientific claim verification dataset are given in Table~\ref{data:stats}. 

\begin{figure}[ht]
    \centering
    \includegraphics[width=0.75\columnwidth]{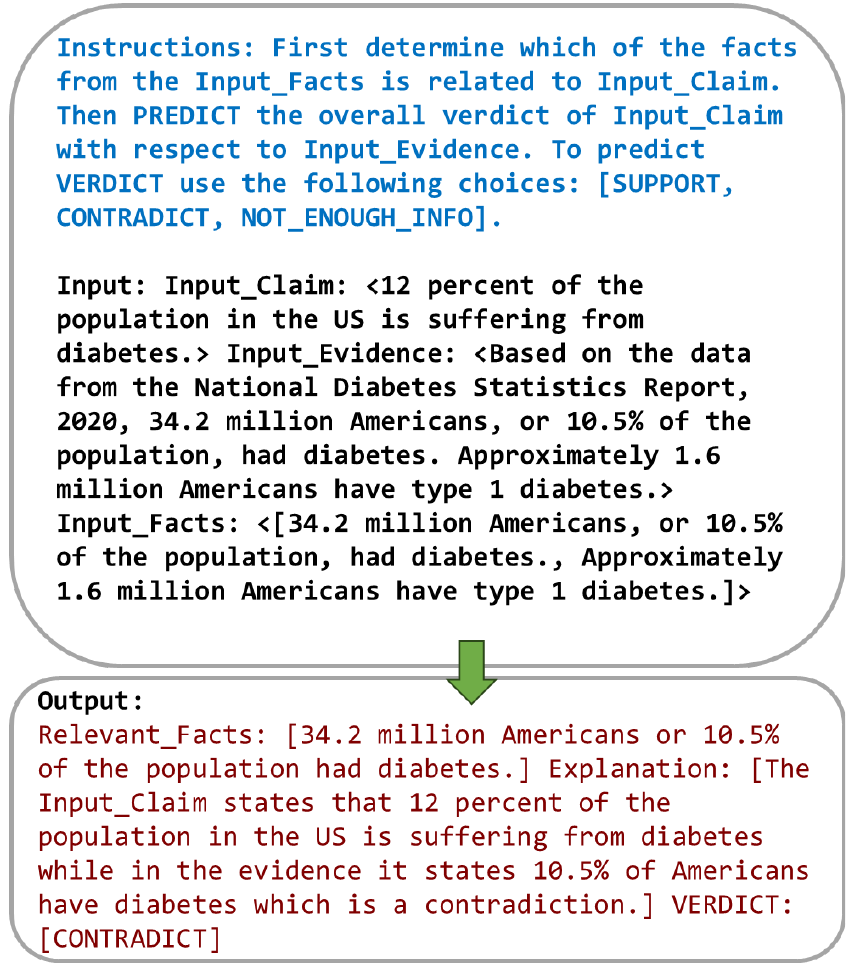}
    \caption{Example of Aug\modelname prompting strategy.}
    \label{fig:a_aug}
\end{figure}
\begin{figure}[ht]
    \centering
    \includegraphics[width=0.75\columnwidth]{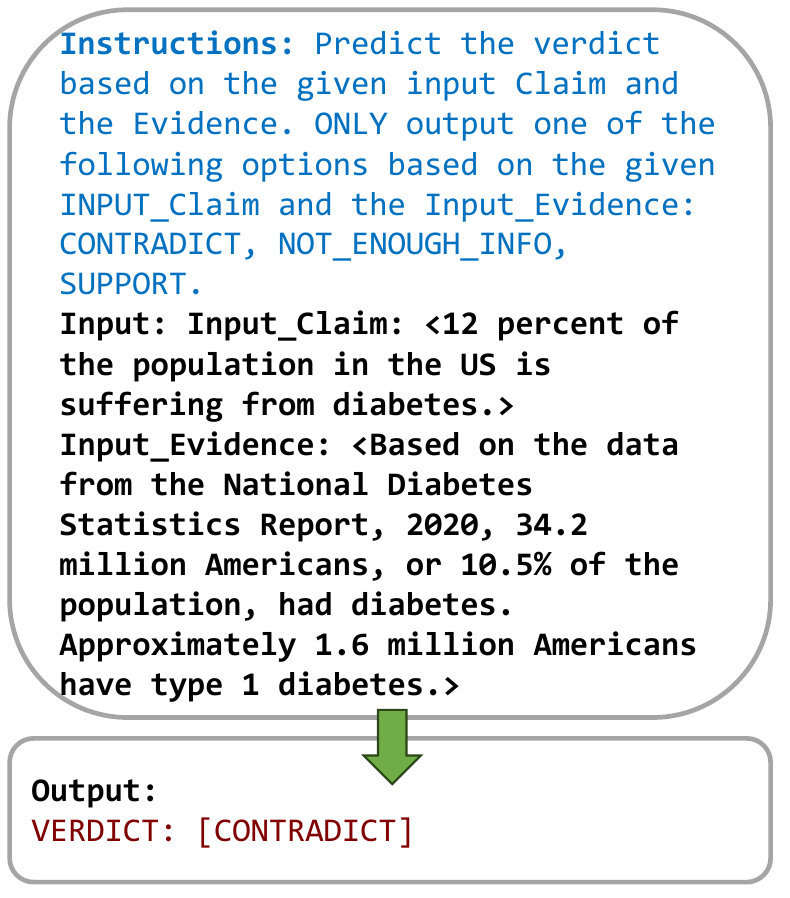}
    \caption{Example of Vanilla prompting strategy.}
    \label{fig:a_van}
\end{figure}
\begin{figure}[h!]
    \centering
    \includegraphics[width=0.75\columnwidth]{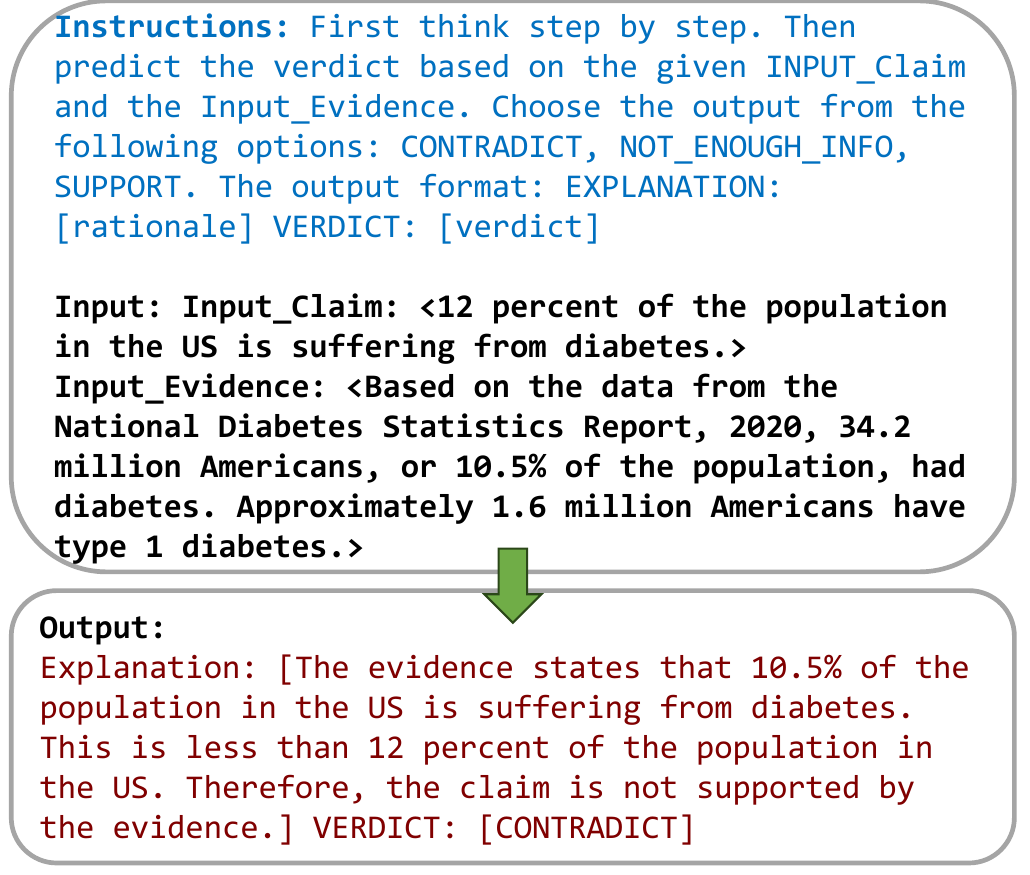}
    \caption{Example of CoT prompting strategy.}
    \label{fig:a_cot}
\end{figure}
\begin{figure}[ht]
    \centering
    \includegraphics[width=0.85\columnwidth]{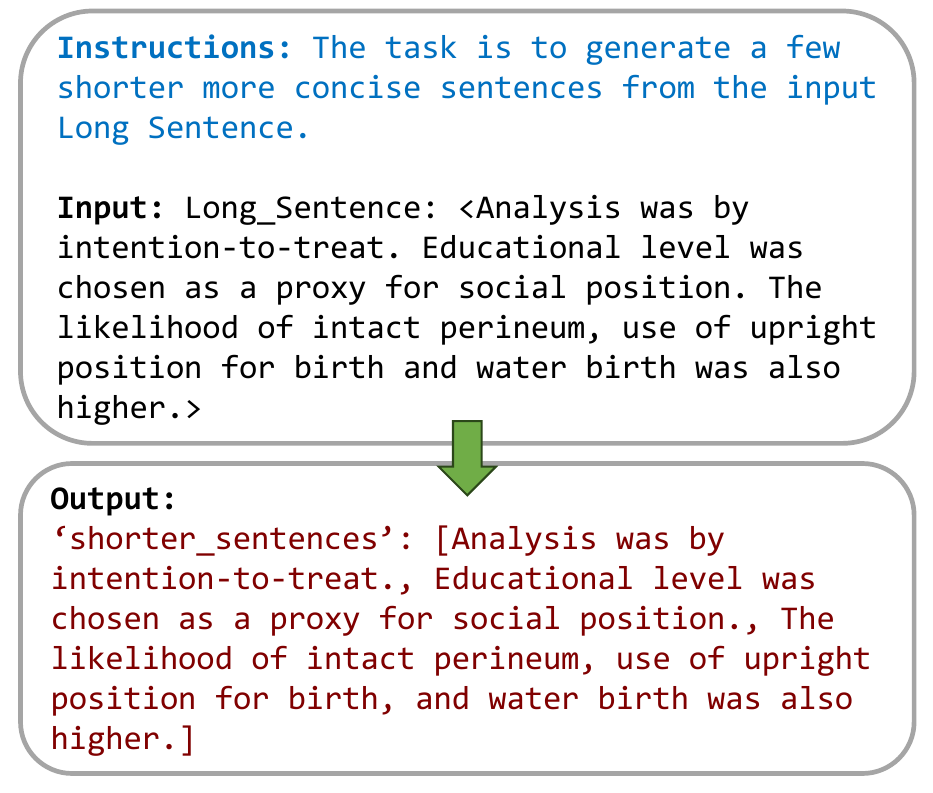}
    \caption{Example of the prompting method used to directly extract short sentences from evidence.}
    \label{fig:a_direct}
\end{figure}
\section{Details of all the Prompting Strategies used in the experiments}~\label{a:prompts}
\subsection{Aug\modelname Prompting Strategy}~\label{a_aug}
Figure~\ref{fig:a_aug} demonstrates the prompt instructions used in this strategy with an example of input and output. First LLMs are prompted to extract the relevant facts from the input facts and then predict the verdict.

\subsection{Vanilla Prompting Strategy}~\label{a_vanilla}
Figure ~\ref{fig:a_van} provides an example of the Vanilla prompting method.

\subsection{CoT Prompting Strategy}~\label{a_cot} 
Figure ~\ref{fig:a_cot} provides an example of the CoT prompting method.

\subsection{Direct Prompting Strategy}~\label{app:direct}

Figure ~\ref{fig:a_direct} provides an example of the prompting method used to directly extract the short sentences along with 5 few shot examples concatenated to the prompt.

 \begin{table}[ht]
 \resizebox{\columnwidth}{!}{
\centering
\begin{tabular}{lcccccc}
\toprule
Base LLM & \multicolumn{3}{c}{Support} & \multicolumn{3}{c}{Contradict} \\
 & F & E & C & F & E & C \\
\midrule

Vicuna-13B & 73.3 & 80.0 & 73.3 & \textbf{90.0} & 73.3 & 70.0 \\
GPT-3.5 & 86.3 & 86.3 & \textbf{81.8} & 70.2 & 59.0 & \textbf{86.0} \\
Mistral-7B & \textbf{83.3} & \textbf{91.0} & {78.1} & 85.2 & \textbf{75.8} & 84.9\\
\bottomrule
\end{tabular}
}
\caption{Human Evaluation results for 3 different LLM \modelname generated short facts.}\label{human_eval}
\end{table}
\section{Human Evaluation of the generated short facts using \modelname} \label{a:human}

 We conducted an experiment to assess the quality of generated short sentences using a manual human evaluation. we manually evaluated three criteria: 1) \textbf{faithfulness} (F), determining if the short sentence is entailed by the evidence, 2) \textbf{essentiality} (E), assessing if the generated sentence is crucial for determining the verdict, and 3) \textbf{conciseness} (C), evaluating if the sentence is sufficiently brief given the evidence. Each sentence was labeled as yes or no. We randomly sampled 15 supported claim-evidence pairs and 15 contradicted ones, evaluating only the originally labeled ``important'' short sentences. Each pair could have multiple short sentences, and we reported the average percentage of yes-labeled sentences per pair. The results of this experiment are presented in Table~\ref{human_eval}. These results show that Mistral-7B generates less concise sentences compared to GPT3.5 whereas it generates more essential sentences. We also see that all the LLMs are at least 70$\%$ faithful to the evidence sentences. Overall Mistral-7B generates higher quality short sentences compared to the other LLMs for this task.

\section{LLM Factuality Evaluation for Document Summarization Through \modelname}~\label{a:factuality}

We show that \modelname is versatile and can be applied to tasks beyond claim verification, such as evaluating the factual consistency of LLM-generated document summaries. To conduct this experiment, we transform the task of evaluating factuality in LLM outputs for document summarization into a claim verification problem. In this setup, the original document serves as evidence, and the summary statement is treated as a claim. We then determine if the statement can be inferred from the document. We then generate short related sentences for the document(evidence) given the statement (claim) using \modelname and perform experiments similar to the claim verification task. In this setup, the only difference is in the output verdict. Instead of prompting LLM to output one of the \emph{Supported, Contradicted and NEI} verdicts, we prompt it if the statement can be inferred from the given document. The output should be either \emph{Yes} or \emph{No}. 
\subsection{Factuality Evaluation Dataset}

We conduct experiments using the Factual Inconsistency Benchmark (FIB~\cite{tam2022fib}) dataset, which includes data from the XSum~\cite{narayan2018don} and CNN/DM~\cite{hermann2015teaching} document summarization datasets. Each instance in the FIB dataset contains two summaries, one of which is factually consistent. For our experiments on the CNN/DM dataset, we use 457 documents, each paired with two statements, one factually consistent and the other not. We label these pairs as "Yes" for factually consistent and "No" for factually inconsistent, resulting in a total of 914 document-statement pairs.
\begin{table}[ht]
\centering

\resizebox{0.85\columnwidth}{!}{
\begin{tabular}{@{}llccc@{}}
\toprule

\multicolumn{1}{c}{Metrics} & Prompt &
   F1 &
  Acc &
  Auc  \\ 
  \toprule
\multirow{4}{*}{FlanT5-XXL}  & Vanilla 
 & 44.1 & 44.2& 57.9\\
 & COT 
 & \underline{44.9} &\underline{44.9} & \underline{58.7}\\

 & Direct 
 & 31.6 &33.0 &53.0\\
 & Aug\modelname 
 & 42.6 & 42.7&53.4 \\
 \midrule
 \multirow{4}{*}{Llama2-13B}  & Vanilla 
 & 19.5 &22.9 &45.1\\
 & COT 
 & 48.6 &56.8 &49.2
\\

 & Direct 
 &54.8  & 63.8 &54.2 \\
 & Aug\modelname 
 & \textbf{58.9} & \textbf{75.0}& \underline{58.3}\\
  \midrule

 \multirow{4}{*}{Mistral-7B}  & Vanilla 
 & 31.4 & 32.2&49.7 \\
 & COT 
 & 34.3 &34.9 &52.9 \\

 & Direct 
 & 38.6 &38.6 &47.4\\
 & Aug\modelname 
 &\underline{46.7}  &\underline{48.7} &\textbf{63.4} \\

 \bottomrule
\end{tabular}
}
\caption{\small{Experimental results for factual consistency evaluation using different prompt strategies. Best performance is highlighted in bold. Underlined values represent best performance for the given LLM.}}\label{res:factuality}
\end{table}
\subsection{Baselines} 
We compare Aug\modelname with Vanilla, CoT, and Direct prompting methods and report the results for 3 open source LLMs of Flan-T5-XXL, Llama2-13B, and Mistral-7B. 

\subsection{Metrics} 
We report results for Macro F1 score, Accuracy, and AUC for this binary classification approach.

\subsection{Results} 
The results are reported in Table~\ref{res:factuality}. We observe that best results are achieved when Aug\modelname is used as prompting method for factual consistency evaluation. Overall decomposing the document into smaller sentences seems to be useful for factual consistency detection and  using \modelname for this task shows superior performance which suggest the effectiveness of \modelname and its applications beyond the claim verification task.

\begin{table*}[ht]
\centering
\resizebox{\textwidth}{!}{
\begin{tabular}{@{}l p{0.55\linewidth} p{0.30\linewidth}@{}}
\toprule
\textbf{Strategy} & \textbf{Input} & \textbf{Generated Output} \\ 
\midrule
Vanilla & \small{\textbf{\textcolor{teal}{Claim:}} Blocking the interaction between TDP-43 and respiratory complex I proteins ND3 and ND6 leads to increased TDP-43-induced neuronal loss. \newline 
\textbf{\textcolor{teal}{Evidence:}} The suppression of TDP-43 mitochondrial localization abolishes WT and mutant TDP-43-induced mitochondrial dysfunction and neuronal loss, and improves phenotypes of transgenic mutant TDP-43 mice} & \small{\textbf{Verdict}: Supported \newline 
\textbf{\textcolor{teal}{Ground Truth:}} Contradicted}\\
\midrule

CoT & \small{\textbf{\textcolor{teal}{Claim:}} 1/2000 in UK have abnormal PrP positivity. \newline 
\textbf{\textcolor{teal}{Evidence:}} RESULTS Of the 32,441 appendix samples 16 were positive for abnormal PrP, indicating an overall prevalence of 493 per million population (95$\%$ confidence interval 282 to 801 per million) }\newline
\emph{Let's Think Step by Step}
& \small{\textbf{Explanation:} The evidence states that an overall prevalence of 493 per million population. This is a counter-factual statement that contradicts with the claim that 1/2000 in UK have abnormal PrP positivity.\newline 
\textbf{Verdict:} Contradicted\newline 
\textbf{\textcolor{teal}{Ground Truth:}} Supported}
\\
\midrule
\modelname & \small{\textbf{\textcolor{teal}{Claim:}} Angiotensin converting enzyme inhibitors are associated with increased risk for functional renal insufficiency. \newline 
\textbf{\textcolor{teal}{Evidence:}}  Although ACE inhibitor therapy usually improves renal blood flow (RBF) and sodium excretion rates in CHF and reduces the rate of progressive renal injury in chronic renal disease, its use can also be associated with a syndrome of “functional renal insufficiency” and/or hyperkalemia. Understanding the pathophysiological mechanisms and the common risk factors for ACE inhibitor–induced functional ARF is critical, because preventive strategies for ARF exist, and if effectively used, they may permit use of these compounds in a less restricted fashion.\newline} 
\newline
\textbf{\textcolor{teal}{Facts:}} 1-angiotensin converting enzyme inhibitors can be associated with functional renal insufficiency and or hyperkalemia.\newline 
2- angiotensin converting enzyme inhibitors are linked to functional renal insufficiency through various pathophysiological mechanisms and common risk factors. preventive strategies are crucial in understanding these connections to mitigate the risk.& \small{\textbf{Relevant Facts:} [angiotensin converting enzyme inhibitors can be associated with functional renal insufficiency and or hyperkalemia., angiotensin converting enzyme inhibitors are linked to functional renal insufficiency through various pathophysiological mechanisms and common risk factors. preventive strategies are crucial in understanding these connections to mitigate the risk.] \newline 
\textbf{Explanation:} he input claim states that angiotensin converting enzyme inhibitors are associated with increased risk for functional renal insufficiency, and the evidence provides relevant supporting facts that ACE inhibitor therapy can be associated with a syndrome of ``functional renal insufficiency'' and/or hyperkalemia..\newline
\textbf{Verdict}: Supported\newline
\textbf{\textcolor{teal}{Ground Truth:}} Supported
}\\
\bottomrule
\end{tabular}
}
\caption{\small{Example prompts used for extracting predictions from GPT-3.5 and their corresponding outputs. The examples are drawn from SciFact dev set.}}
\label{prompt_examples}
\end{table*}

\begin{table*}[h!]
\centering
\begin{tabular}{|p{5cm}|p{5cm}|p{5cm}|}
\hline
\textbf{Example} & \textbf{Direct} & \textbf{FactDetect} \\
\hline
\textbf{Claim:} Antiretroviral therapy reduces rates of tuberculosis across a broad range of CD4 strata. \newline \textbf{Evidence:} Antiretroviral therapy is strongly associated with a reduction in the incidence of tuberculosis in all baseline CD4 count categories: (1) less than 200 cells/µl (hazard ratio [HR] 0.35, 95\% CI 0.28 to 0.44). CONCLUSIONS Antiretroviral therapy is strongly associated with a reduction in the incidence of tuberculosis across all CD4 count strata. & 
\begin{itemize}
\item antiretroviral therapy is strongly associated with a reduction in the incidence of tuberculosis.
\item 95$\%$ confidence interval ci 0.07 to 0.36)
\item 200 to 350 cells/µl (hr 0.34
\item antiretroviral therapy is strongly associated with a reduction in the incidence of tuberculosis in all baseline cd4 count categories: less than 200 cells/µl (hazard ratio hr 0.16"
\item conclusions: antiretroviral therapy is strongly associated with a reduction in the incidence of tuberculosis across all cd4 count strata.
\end{itemize} & 
\begin{itemize}
\item antiretroviral therapy reduces the incidence of tuberculosis in all baseline cd4 count categories less than 200 cells/µl, with a hazard ratio of 0.16 (95\% ci: 0.07-0.36)
\end{itemize} \\
\hline
\textbf{Claim:} Autologous transplantation of mesenchymal stem cells causes a higher rate of opportunistic infections than induction therapy with anti-interleukin-2 receptor antibodies. \newline \textbf{Evidence:} Also, during the 1-year follow-up, combined analysis of MSC-treated groups revealed a significantly decreased risk of opportunistic infections than the control group (hazard ratio, 0.42; 95\% CI, 0.20-0.85, P=.02)  ... and better estimated renal function at 1 year. & 
\begin{itemize}

\item during the 1-year follow-up combined analysis of msc-treated groups revealed significantly decreased risk of opportunistic infections than the control group.
\item conclusion among patients undergoing renal transplant
\item the use of autologous mscs compared with anti-il-2 receptor antibody induction therapy resulted in lower incidence of acute rejection \item decreased risk of opportunistic infection and better estimated renal function at 1 year. 

\end{itemize} & 
\begin{itemize}
\item autologous transplantation of mesenchymal stem cells increases the risk of opportunistic infections compared to induction therapy with anti-interleukin-2 receptor antibodies, as shown by a significantly decreased risk of opportunistic infections in the control group compared to the msct-treated groups.
\item autologous transplantation of mesenchymal stem cells increases the risk of opportunistic infections compared to induction therapy with anti-interleukin-2 receptor antibodies by a hazard ratio of 0.42 (95$\%$ ci 0.20-0.85), with a p-value of 0.02.
\end{itemize} \\
\hline
\end{tabular}
\caption{\small{Example of the \modelname generated short facts and Direct approach generated short facts for 2 examples from SciFact Dev set.}}
\end{table*}
\end{document}